\documentclass[runningheads]{llncs}
\usepackage{cite}
\usepackage{amsmath,amssymb,amsfonts}
\usepackage{algorithmic}
\usepackage{textcomp}
\usepackage{xcolor}
\usepackage{comment}
\usepackage{hyperref}
\usepackage{tikz}
\usepackage{subfig}
\usepackage{pgfplots}
\usepackage{multirow}
\usepackage[linesnumbered,ruled,vlined]{algorithm2e}
\usepackage{todonotes}
\usepackage[T1]{fontenc}
\usepackage{graphicx}
\begin{document}
\title{Beyond Yes or No: Predictive Compliance Monitoring Approaches for Quantifying the Magnitude of Compliance Violations}
\titlerunning{Predictive Compliance Monitoring}
\author{Qian Chen\orcidID{0009-0004-1326-8898}\inst{1} \and 
Stefanie Rinderle-Ma\orcidID{0000-0001-5656-6108}\inst{1} \and 
Lijie Wen\orcidID{0000-0003-0358-3160}\inst{2}} %
\authorrunning{Q. Chen et al.}
\institute{Technical University of Munich, Garching, Germany\\
\email{\{qian.chen,stefanie.rinderle-ma\}@tum.de} \and
Tsinghua University, Beijing, China\\
\email{wenlj@tsinghua.edu.cn}}
\maketitle              % typeset the header of the contribution
\begin{abstract}
Most existing process compliance monitoring approaches detect compliance violations in an ex post manner. Only predicate prediction focuses on predicting them. However, predicate prediction provides a binary yes/no notion of compliance, lacking the ability to measure to which extent an ongoing process instance deviates from the desired state as specified in constraints. Here, being able to quantify the magnitude of violation would provide organizations with deeper insights into their operational performance, enabling informed decision making to reduce or mitigate the risk of non-compliance. Thus, we propose two predictive compliance monitoring approaches to close this research gap. The first approach reformulates the binary classification problem as a hybrid task that considers both classification and regression, while the second employs a multi-task learning method to explicitly predict the compliance status and the magnitude of violation for deviant cases simultaneously.
In this work, we focus on temporal constraints as they are significant in almost any application domain, e.g., health care. The evaluation on synthetic and real-world event logs demonstrates that our approaches are capable of quantifying the magnitude of violations while maintaining comparable performance for compliance predictions achieved by state-of-the-art approaches. 

\end{abstract}
\section{Introduction}\label{sec:introduction}

Process compliance is a pivotal component in information systems research \cite{DBLP:conf/caise/AbdullahSI10} and focuses on ensuring that processes adhere to compliance constraints that are imposed on them. Compliance constraints may stem from legislative and regulatory documents, standards and codes of practice, and business rules or contracts \cite{DBLP:journals/kais/HashmiGLW18}, might change frequently (e.g., financial regulations change every $12$ minutes\footnote{\url{https://thefinanser.com/2017/01/bank-regulations-change-every-12-minutes}}), and non-compliance might result in the risk of reputational and financial loss and even criminal penalties \cite{e5d28f956f7f4e3984a5ff133c5efaa3}. 

Several surveys on challenges in compliance management in (process-aware) information systems exist, e.g., \cite{DBLP:conf/caise/AbdullahSI10,DBLP:journals/is/LyMMRA15,DBLP:journals/isf/LyRGD12,DBLP:journals/is/RinderleMaWB23}. A first dimension of challenges arises from the process life cycle phase when compliance verification is supposed to take place, i.e., design time, runtime and ex post. Compliance by design \cite{DBLP:journals/flap/TosattoGBO19} verifies the compliance of a process model with the set of compliance constraints. During runtime, ongoing process instances which are generated from the process model are monitored to detect or even predict compliance violations (i.e., compliance monitoring) \cite{DBLP:journals/is/LyMMRA15}. Afterwards, the set of executed traces (an event log) are compared to a relevant process model to assess their conformity (i.e., conformance checking) \cite{carmona2018conformance}. Despite the importance of design and ex post compliance checking, (predictive) compliance monitoring enables business organizations to proactively detect and manage possible compliance violations during process execution \cite{DBLP:journals/is/LyMMRA15}. However, only few existing approaches target the prediction of violations. Hence, in this work, we focus on the sub-field of compliance monitoring, i.e., \textit{predictive compliance monitoring (PCM)} \cite{DBLP:journals/is/RinderleMaWB23}.

Another challenge is the lack of sufficient \textsl{ability to quantify the degree of compliance} defined as a compliance monitoring functionality (CMF10) in \cite{DBLP:journals/is/LyMMRA15}. To predict compliance violations at runtime, most approaches \cite{DBLP:conf/caise/MaggiFDG14,DBLP:journals/tsc/Francescomarino19,DBLP:journals/access/ZhangL20q} offer binary yes/no notions of compliance. \textsl{Outcome-oriented predictive process monitoring} \cite{DBLP:journals/tkdd/TeinemaaDRM19} and \textsl{predicate prediction} \cite{DBLP:conf/caise/MaggiFDG14} aim to classify each running case based on a predefined set of \textit{categorical} outcomes, e.g., will an order be delivered on time (desired outcome) or not (undesired outcome)? However, these approaches lack the ability to quantify the extent to which an ongoing process case will deviate from the desired state. The ability to quantify the magnitude of violation is crucial, particularly for prescriptive process monitoring techniques. These techniques aim to recommend or prescribe actions to prevent undesired outcomes by minimizing a cost function of interventions. Such cost function considers factors like the probability of an undesired outcome occurring, the cost of executing interventions, and the timing of those interventions (i.e., their mitigation effectiveness) \cite{DBLP:journals/kais/Fahrenkrog-Petersen22}. Including the magnitude of violation as a key factor in the cost function is essential, as severe deviations may require more substantial and costly interventions compared to mild ones. In addition, quantifying the magnitude of violation supports process participants in making risk-informed decisions to reduce the likelihood or severity of process faults \cite{DBLP:journals/dss/ConfortiLRAH15}. For instance, severe violations can be prioritized for intervention to mitigate their impact, while cases with mild extent of violation may be addressed first, as they are often easier and less costly to resolve compared to severe ones.

To address these challenges, we propose two PCM approaches which extend existing predicate prediction methods with the ability to quantify the magnitude of violations. Here, the magnitude of violation corresponds to the degree of compliance \cite{DBLP:journals/is/LyMMRA15} as utilized in literature. The hybrid approach reformulates the binary classification problem into a regression task, implicitly preserving the nature of binary classification with a focus on violation predictions. The multi-task learning (MTL) approach handles both tasks—compliance prediction and magnitude quantification—explicitly. MTL is particularly suitable for our problem because these two tasks are inherently interdependent. Predicting the compliance status provides a broad categorization by separating normal and deviant cases, while quantifying the magnitude of violations focuses specifically on deviant cases, offering a deeper analysis. In this work, we focus on quantifying the magnitude of violation in terms of time, because meeting temporal restrictions (e.g., strict deadlines) is deemed vital in many application domains such as the logistics and aviation industry \cite{cheikhrouhou2015temporal}. Moreover, timestamps are available in all event logs.

We evaluated our approaches based on synthetic and real-life event logs. The results show that our proposed approach, MTL in particular, is able to quantify the magnitude of violation while maintaining comparable compliance prediction performance achieved by the state-of-the-art predicate prediction approaches.

Section \ref{sec:problem} presents a running example. Our proposed predictive compliance monitoring approaches are introduced in Sect. \ref{sec:approach} and evaluated in Sect. \ref{sec:evaluation}. Related work is discussed in Sect. \ref{sec:rel_work}, followed by the conclusion in Sect. \ref{sec:conclusion}.

\section{Problem Statement}\label{sec:problem} 

\begin{figure}
    \centering
    \includegraphics[width=\columnwidth]{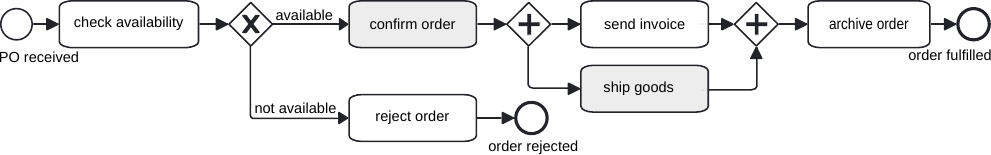}
     \caption{Order-to-Cash Process Model}
    \label{fig:running_example}
\end{figure}

We illustrate the problem based on an Order-to-Cash (O2C) process (cf. Fig. \ref{fig:running_example}) as the running example. The process is triggered when a purchase order (PO) is received. An associated temporal constraint $\varphi_t$ is introduced into the process model that imposes a maximum temporal distance of $24$ hours between tasks \texttt{ship goods} and \texttt{confirm order}.

\begin{figure}
    \centering
    \includegraphics[width=\columnwidth]{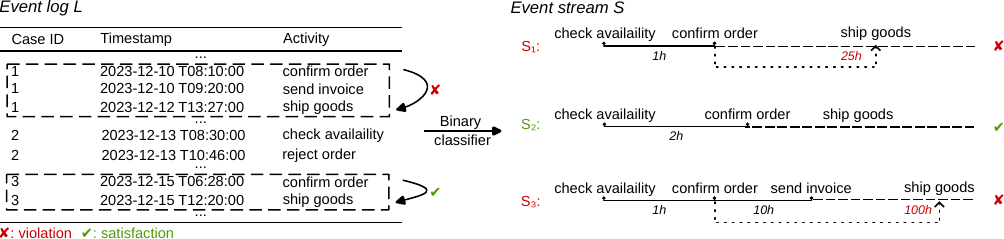}
     \caption{Predicate Prediction}
    \label{fig:predicate_prediction}
\end{figure}

To predict compliance states of running cases w.r.t. the associated temporal constraint $\varphi_t$, existing predicate prediction approaches \cite{DBLP:conf/caise/MaggiFDG14,DBLP:journals/tsc/Francescomarino19,DBLP:journals/access/ZhangL20q} classify them either into satisfaction or violation via a binary classifier. Consider the event log that has been derived from the running example in Fig. \ref{fig:predicate_prediction}. Predicate prediction approaches \cite{DBLP:conf/caise/MaggiFDG14} or outcome-oriented predictive process monitoring (outcome-oriented PPM) \cite{DBLP:journals/tkdd/TeinemaaDRM19} first assign a boolean outcome to every case in the log according to $\varphi_t$, i.e., positive if the case violated $\varphi_t$ or negative if it satisfied with $\varphi_t$. The binary classifier is then trained based on the labeled event log and applied to evolving event streams, aiming to predict which of the running case will and will not violate $\varphi_t$. 

By contrast, this work does not aim only for a yes/no answer, but also at measuring the magnitude of violation that enables business organizations to gain a deeper insight into the operational performance. Consider streams $s_1$ and $s_3$ in Fig. \ref{fig:predicate_prediction} and assume a predicted duration of $25$ and $100$ hours between tasks \texttt{confirm order} and \texttt{ship goods} for $s_1$ and for $s_3$, respectively. Existing predicate prediction approaches would classify both as non-compliant, but $s_3$ exhibits a more significant extent of deviation. Differentiating non-compliance levels of process executions can support an organization in making informed decisions to reduce or mitigate the risk of non-compliance. Potentially non-compliant process instances with a ``mild degree'' of violation such as $s_1$ with anticipated time of $25$ hours, for example, can then be prioritized in light of the limited resources at disposal in order to \textsl{avoid risk}. Moreover, the magnitude of violation is considered as an important factor when predicting risks (\textsl{severity} of risks \cite{DBLP:journals/dss/ConfortiLRAH15}) and prescribing when and how to intervene (\textsl{intervention effectiveness} \cite{DBLP:journals/kais/Fahrenkrog-Petersen22}), with the aim to reduce the cost of non-compliance.

\section{Approach}
\label{sec:approach}

Existing predicate prediction approaches \cite{DBLP:conf/caise/MaggiFDG14,DBLP:journals/tsc/Francescomarino19,DBLP:journals/access/ZhangL20q,DBLP:journals/tkdd/TeinemaaDRM19} focus on predicting compliance violations in a binary yes/no notion of compliance. Hence, they lack the capability to quantify the magnitudes of violations for deviant cases. By contrast, simply framing the problem as a regression task to predict the extent of deviation which can be either positive or negative signaling satisfied or violated cases cannot emphasize violation predictions. To address this research gap we combine the prediction of the occurrence of compliance violations (classification) with predicting their magnitude (regression) for ongoing cases. To this end, we propose hybrid and multi-task learning (MTL) approaches. We first introduce the pipeline of existing predicate prediction approaches to describe the procedure for compliance predictions. The proposed approaches are presented afterwards.

\subsection{Predicate Prediction Pipeline}\label{sec:predicate prediction}

\begin{figure}
    \centering
    \includegraphics[width=\textwidth]{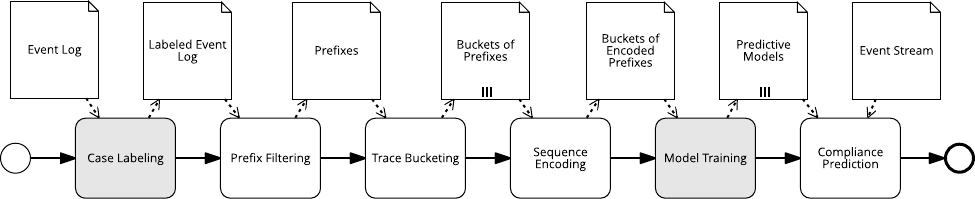}
     \caption{Pipeline of Predicate Prediction Approaches}
    \label{fig:approach_pipeline}
\end{figure}

Figure \ref{fig:approach_pipeline} shows the pipeline of the state-of-the-art methods for compliance prediction. The first step is to label each case in the event log based on the associate constraint, and prefixes are generated and filtered based on the different requirements, e.g, only include prefixes up to a maximum length. Next, all retained prefixes can be divided into several buckets based on the length of prefixes or various clustering algorithms, and different classifiers are then trained for each such buckets. If all prefixes are considered into the same bucket, then a single classifier is trained based on all prefixes. For prefixes in each bucket, we apply different encoding techniques for attributes of the event within the trace. The most common approach is to one-hot encode categorical variables and apply normalization to numerical attributes. Finally, the predictive model is trained based on the encoded feature vectors and applied to an ongoing case to determine whether it's compliant or not. In the following, we focus on the most critical steps—case labeling and model training—as highlighted in Fig.~\ref{fig:approach_pipeline}, since these steps represent the key distinctions between conventional predicate prediction approaches and our proposed methods.

\noindent\textbf{Case Labeling} is conducted for each case in the event log based on the corresponding temporal compliance constraint. An event log $L:=\{\sigma_1, ..., \sigma_k\}$ comprises a set of completed traces $\sigma_i$ where $\sigma_i:=\langle e_1,e_2,...,e_n\rangle$ is defined as a sequence of events $e_j$. The prefix $hd^k(\sigma_i):=\langle e_1,e_2,...,e_k\rangle$ is defined as a partial trace of $\sigma_i$ where $k\leq |\sigma_i|$. An event $e_j$, in turn, is defined as a tuple $(a,c,t,d_1,...,d_m)$ where $a$ is an activity type, $c$ is the case id, $t$ is the timestamp of execution, and $d_1,..,d_m$ are other attributes of this event.
A temporal constraint $\varphi_t$ specifies both the control-flow and temporal requirements each process execution must comply with. The control-flow patterns specify the occurrences or ordering of activities, such as existence and eventually follows relations. Time patterns, on the other hand, impose restrictions on the (e.g., minimum or maximum) temporal distance between the execution of activities.

\begin{figure}
    \centering
    \includegraphics[width=0.8\columnwidth]{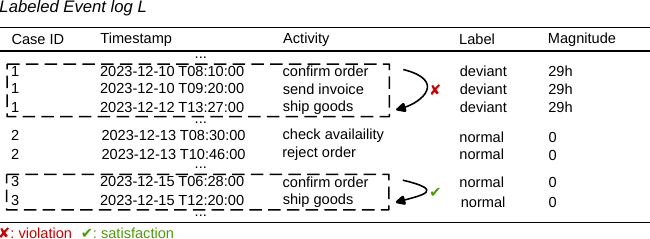}
     \caption{Labeled Event Log}
    \label{fig:labeled_log}
\end{figure}

The labeling function for each case w.r.t. the associated constraint $\varphi_t$ is defined as follows:
\[
y_{\text{binary}} (\sigma) =
\begin{cases} 
\text{deviant} \rightarrow{1} & \text{if $\varphi_t$ violated in $\sigma$} \\
\text{normal} \rightarrow{0} & \text{otherwise}
\end{cases}
\]
Specifically, for a given temporal constraint $\varphi_t$, a case is labeled as deviant (encoded as 1) if it violates the constraint; otherwise, it is labeled as normal (encoded as 0). Consider the running example as introduced in Sect.~\ref{sec:problem} with $\varphi_t$: ``Goods must be shipped no later than 24 hours after order confirmation''. The labeled event log after applying $\varphi_t$ to every case in the log is shown in Fig.~\ref{fig:labeled_log}. The magnitude column will be explained later in our proposed approaches.

\noindent\textbf{Model Training} takes as input the encoded feature vectors from the prefixes log and
predicts their corresponding labels. In order to determine whether a running process execution is going to comply with or violate the associated temporal constraint, a simple binary classifier, e.g., decision tree (DT) or support vector machines (SVM), is often adopted as the predictive model in existing predicate prediction approaches. For a binary classification problem, we target at minimizing the binary cross-entropy loss (BCE) between the true labels $y_i$ (either 0 or 1) and the predicted values $\hat{y}_i$ (i.e., estimated probability within $[0,1]$) for all (i.e., $n$) samples. The loss function is defined as follows:
\begin{equation}
\mathcal{L}_{\text{BCE}} = -\sum_{i=1}^{n} y_i \log(\hat{y}_i)
\label{eq:CE}
\end{equation}

\subsection{Hybrid Approach}\label{sec:hybrid approach}%\todo{describe the idea of hybrid method from \cite{junior2023hybrid} first}
In order to predict the compliance of running cases and to further assess the degree of such violations, a pure classification or regression method is not capable of achieving both goals. Thus, we present a hybrid approach which is inspired by \cite{junior2023hybrid} in grid distribution networks. It targets at predicting the occurrence of grid constraint violations as well as the amplitude of such violation, aiming to identify and prevent grid constraint violations for the stable and efficient operation of systems.

\noindent\textbf{Case Labeling} in this approach is different from the conventional predicate prediction methods as introduced in Sect.~\ref{sec:predicate prediction}. We follow up on the idea from \cite{junior2023hybrid} to assign the magnitude of violation for each case in the log, extending the binary classification problem by incorporating regression to capture the degree of violations. In particular, the hybrid approach assigns a positive value to deviant cases to reflect the magnitude of violation, while normal cases are consistently labeled as 0. For deviant cases that breach temporal constraints in the control-flow dimension, we use the case duration (i.e., the time from the start to the end of the case) as the magnitude of violation to highlight the severity of non-compliance. By doing so, it maintains consistency with the binary classification method for normal cases while offering a more nuanced evaluation of violations.

The labeling function adopted in the hybrid approach is defined as follows:
\[
y_{\text{hybrid}} (\sigma) =
\begin{cases} 
\text{magnitude of violation} & \text{if $\varphi_t$ violated in $\sigma$} \\
0 & \text{otherwise}
\end{cases}
\]
Consider the running example again with $\varphi_t$ imposing a maximum temporal restriction of 24 hours between \texttt{confirm order} and \texttt{ship goods}. Applying $y_{\text{hybrid}}$ for each case in the log results in the magnitude column (cf. Fig,~\ref{fig:labeled_log}) as the target values. For example, the duration between these two events in $\sigma_1$ is roughly 53 hours based on the corresponding timestamps, then the magnitude of violation is calculated as $53-24=29$ hours.

\noindent\textbf{Model Training} takes the same feature vectors as input, but predicts the magnitudes of violations instead of class labels. To enable compliance predictions, we  convert the continuous predictions to binary values according to the hybrid labeling function $y_{\text{hybrid}}$. Specifically, cases are classified as deviant if their predicted values $>0$, otherwise, they are considered normal. For the regression task, we aim to minimize the squared difference between the ground truth values $y_i$ and predicted values $\hat{y}_i$ by using the mean squared error (MSE) as the loss function:
\begin{equation}
\mathcal{L}_{\text{MSE}} = \frac{1}{n} \sum_{i=1}^{n} \left( y_i - \hat{y}_i \right)^2    
\label{eq:MSE}
\end{equation}

\subsection{Multi-task Learning}
The multi-task learning (MTL) approach is particularly suited to our problem as it allows the model to jointly learn compliance prediction (binary classification) and magnitude quantification (regression). Both tasks are closely related: the classification task identifies whether a case is compliant or deviant, while the regression task quantifies the magnitude of violation for deviant cases. By sharing knowledge across tasks, MTL exploits their interdependence, enabling the model to learn richer feature representations that improve both tasks. By doing so, this framework provides both a high-level understanding of compliance and detailed insights into the magnitude of violations, enhancing overall predictive performance. Thus, the MTL approach incorporates both the predicate prediction approach (cf. Sect.~\ref{sec:predicate prediction}) and the hybrid technique (cf. Sect.~\ref{sec:hybrid approach}) to predict compliance states and the magnitude of violation simultaneously.

\noindent\textbf{Case Labeling} for MTL approach combines labeling functions from predicate prediction ($y_{\text{binary}}$) and hybrid approach ($y_{\text{hybrid}}$) to assign labels and the magnitudes of violations for each case, resulting in a labeled event log with both label and magnitude columns as shown in Fig.~\ref{fig:labeled_log}. Thus, the integrated labeling function is updated as follows:
\[
y_{\text{mtl}} (\sigma) =
\begin{cases} 
y_{\text{binary}} (\sigma) \\
y_{\text{hybrid}} (\sigma)
\end{cases}
\]

\begin{figure}
    \centering
    \includegraphics[width=0.45\columnwidth]{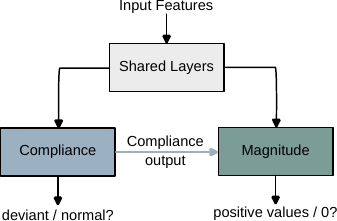}
     \caption{Model Architecture for MTL}
    \label{fig:model_architecture}
\end{figure}

\noindent\textbf{Model Training} in the MTL framework addresses both classification and regression tasks simultaneously. As illustrated in Fig.\ref{fig:model_architecture}, 
the compliance prediction and magnitude of violation quantification tasks share the same layers, because they share common features and representations. The compliance prediction task predicts whether a case is deviant or normal, leveraging the binary cross-entropy loss function introduced in Sect.~\ref{sec:predicate prediction}. For the magnitude of violation prediction, unlike the hybrid approach, the model incorporates the output of the compliance prediction task as an additional input, alongside the shared features. This integration is crucial because the magnitude of violation is only relevant for deviant cases, making the classification output a valuable signal for the regression task. 
Afterwards, it outputs either a positive value, representing the magnitude of violation, or 0, indicating satisfaction for the given case.

To train the MTL model, we minimize the total loss, which combines the losses from both compliance prediction (cf. Equation \ref{eq:CE}) and violation magnitude quantification (cf. Equation \ref{eq:MSE}):
\begin{equation}
\mathcal{L} = \mathcal{L}_{\text{BCE}} + \mathcal{L}_{\text{MSE}}    
\end{equation}

\section{Evaluation}\label{sec:evaluation}
In this section, we demonstrate the efficacy of our proposed approaches by comparing them with the conventional binary predicate prediction approach (baseline). We employ the XGBoost classifier as one of the predictive models, as it has been shown to outperform other traditional machine learning models in the benchmark study of outcome-oriented PPM approaches \cite{DBLP:journals/tkdd/TeinemaaDRM19}. Additionally, we incorporate a deep learning based predictive model---the Attention-based Bidirectional LSTM Neural Network (Att-Bi-LSTM)---as the second baseline due to its superior performance compared to traditional machine learning methods \cite{DBLP:conf/icws/WangYLS19}. Moreover, the MTL approach is well-suited for deep learning models as it requires flexible adaptation of the model architecture. This section introduces the experimental setup, including datasets and evaluation metrics, followed by data preprocessing and model adaptation. The experiment results are provided afterwards. We use the code provided by \cite{DBLP:journals/tkdd/TeinemaaDRM19} for experiments with XGBoost. The implementation for Att-Bi-LSTM (source code is not provided in \cite{DBLP:conf/icws/WangYLS19}) is available at \url{https://github.com/Qian915/predictive-compliance-monitoring}.

\subsection{Experimental Setup}
All experiments are conducted on a synthetic data set and two real-life event logs with temporal constraints involved.

\noindent\textbf{Datasets.} The synthetic data set o2c is generated from the Order-to-Cash process (cf. Fig. \ref{fig:running_example}) using the Cloud Process Execution Engine\footnote{\url{https://cpee.org/}} (CPEE) \cite{DBLP:journals/corr/abs-2208-12214}. For real-life event logs, the sepsis event log \footnote{\url{https://data.4tu.nl/articles/dataset/Sepsis_Cases_-_Event_Log/12707639}} records patient pathways in a hospital for sepsis treatment, and bpic2012w \footnote{\url{https://data.4tu.nl/articles/BPI_Challenge_2012/12689204}} deals with a sub-process (\textsl{work item}) within a loan application procedure. 

Table \ref{tab:datasets} presents the characteristics of all datasets, and the corresponding temporal constraints are presented as follows:

\noindent o2c\_1: ``Goods must be shipped no later than 24 hours after order confirmation.''

\noindent sepsis\_1:  ``Patients should be administered antibiotics within one hour after ER Sepsis Triage''

\noindent sepsis\_2:  ``Lactic Acid measurements should be performed within three hours from ER Sepsis Triage.''

\noindent bpic2012w\_1: ``A scheduled validation should start within 2 days.''

\noindent bpic2012w\_2: ``The validation must be done in at most 20 minutes.''

\begin{table}[htb!]
\centering
\caption{Characteristics of Datasets}\label{tab:datasets}
\begin{tabular}{c c c c c c c c}
\hline
\multirow{2}{*}{Dataset} & \multirow{2}{*}{\#traces} & \multirow{2}{*}{\#event} & \multicolumn{3}{c}{Length}  & \multirow{2}{*}{pos class} & \multirow{2}{*}{act vio} \\
\cline{4-6}
& & classes & min. & avg. & max. & ratio & ratio \\
\hline
o2c\_1 & 998 & 6 & 1 & 2.53 & 3 & 0.41 & 0.0 \\
sepsis\_1 & 782 & 15 & 1 & 7.19 & 61 & 0.63 & 0.2 \\
sepsis\_2 & 782 & 16 & 3 & 7.08 & 39 & 0.25 & 0.94 \\
bpic2012w\_1 & 9658 & 15 & 2 & 13.17 & 93 & 0.20 & 0.02 \\
bpic2012w\_2 & 9658 & 17 & 2 & 13.61 & 93 & 0.15 & 0.0 \\
\hline
\end{tabular}
\end{table}

\noindent\textbf{Metrics.} 
For the binary compliance prediction task, we choose \textsl{area under the ROC curve (AUC)} instead of the most commonly used metric \textsl{accuracy}, because it remains unbiased even on imbalanced datasets and is threshold-independent \cite{DBLP:journals/pr/Bradley97}. AUC represents the probability that a positive instance is ranked higher than a negative one. For the regression problem of predicting the magnitude of violations, \textsl{Mean Absolute Error (MAE)} is used to evaluate the average prediction error by calculating the mean of the absolute differences between predicted and actual values. Since the binary classification approach does not provide magnitude predictions, we instead evaluate a trivial baseline approach by using the average of the actual values as predictions and calculating the mean of the absolute differences between this average and the actual values. The MAE values for our approaches and the baseline are presented as follows:

\begin{equation}
 MAE = \frac{1}{n} \sum_{i=1}^{n} \left| y_i - \hat{y}_i \right|   
\end{equation}

\begin{equation}
MAE_{\text{baseline}} = \frac{1}{n} \sum_{i=1}^{n} \left| y_i - \bar{y} \right|   
\end{equation}
Here, $y_i$ and $\hat{y}_i$ are the true and predicted magnitude of violation, $\bar{y}$ is the mean of true values, and $n$ is the total number of test prefixes. The baseline MAE sets a reference point to assess whether our approach offers a meaningful improvement.
%Here, $y_i$ and $\hat{y}_i$ denotes the true and predicted magnitude of violations in the test set, $\bar{y}$ is the mean of true values, and $n$ is the total number of prefixes in the test set. This baseline MAE serves as a reference point, setting a standard for minimum performance and enabling us to determine whether our approach provides a meaningful improvement over this baseline.

\subsection{Data Preprocessing and Model Adaptation}
All event logs are first labeled based on the corresponding labeling function as introduced in Sect.~\ref{sec:approach}. For each dataset, we remove incomplete cases and include all attributes available in the event log as input features. In addition, time-related features, such as \textsl{month}, \textsl{weekday} and \textsl{hour}, are extracted based on the timestamp attribute as we also target at quantifying the magnitude of temporal violations. All traces are cut before the relevant events derived from the temporal constraint happen to allow early predictions, i.e., cut traces exactly before the corresponding class labels are known and irreversible. In this work, we do not consider the multiple occurrences of constraint-related events, thus, traces cutting can be applied even to cases with eventually follows relation involved. We keep the 80\%-20\% train-test-split ratio for model training and evaluation. Prefixes are extracted from the labeled event log and filtered based on a predefined maximum prefix length. For each dataset, the maximum prefix length is determined as the case length corresponding to the $90th$ percentile of positive cases as adopted in \cite{DBLP:journals/tkdd/TeinemaaDRM19}. For our experiments, we train the predictive model based on all retained prefixes, namely prefixes are not divided into several buckets. 

For data preprocessing, the main difference between XGBoost and Att-Bi-LSTM lies in how events are encoded. For XGBoost, the optimal encoding technique is aggregation: categorical attributes are represented by their frequency of occurrences, while numerical attributes are denoted with the mean and standard deviation \cite{DBLP:journals/tkdd/TeinemaaDRM19}. By contrast, Att-Bi-LSTM uses one-hot encoding for categorical features and applies normalization to numerical values \cite{DBLP:conf/icws/WangYLS19}.

\begin{figure}
    \centering
    \includegraphics[width=0.5\columnwidth]{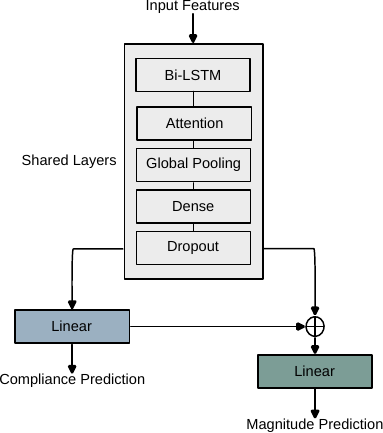}
     \caption{Model Architecture for Att-Bi-LSTM}
    \label{fig:att-bi-lstm_model_architecture}
\end{figure}

\noindent\textbf{Model Adaptation.} Traditional machine learning models do not support model adaptation. Therefore, for XGBoost, we modify only the loss function as described in  Sect.~\ref{sec:approach}, based on the corresponding approach. As for deep learning models, we adapt the Att-Bi-LSTM model architecture to enable multi-task predictions (cf. Fig.\ref{fig:att-bi-lstm_model_architecture}). Input features are first processed through a bidirectional LSTM layer, followed by an attention layer that identifies the most critical parts of inputs. These outputs are then passed through a global average pooling layer, a dense layer, and a dropout layer.

For the binary predicate prediction approach (baseline), all features are fed into a linear layer to predict whether the temporal constraint will be satisfied or violated in the trace. The hybrid approach also follows this structure to quantify the magnitude of violation. However, in the multi-task learning approach, compliance and magnitude predictions are performed simultaneously. In particular, for the magnitude prediction task, it takes as input the shared features concatenated with the output from the compliance prediction task representing captured patterns for separating satisfaction and violation.

Furthermore, we optimize hyperparameters using the Tree-structured Parzen Estimator (TPE) algorithm \cite{DBLP:conf/nips/BergstraBBK11}. Additionally, a 3-fold cross-validation is conducted for each hyperparameter configuration.

\subsection{Results}
We report experimental results with XGBoost (cf. Tab. \ref{tab:results_xgboost}) and Att-Bi-LSTM (cf. Tab.~\ref{tab:results_lstm}) to assess the performance of compliance prediction (AUC) and magnitude quantification (MAE in days). 

\noindent\textbf{XGBoost.}
For traditional machine learning models, such as XGBoost (cf. Tab. \ref{tab:results_xgboost}) in this study, our proposed approaches outperform the baseline in predicting the magnitude of violations, achieving lower MAE values across nearly all datasets. The only exception is the hybrid approach on the sepsis datasets, where it underperforms the baseline with higher MAE values. This underperformance can be attributed to the high proportion of deviant cases which violate the temporal constraints on the control-flow dimension (cf. activity violation ratio in Tab.~\ref{tab:datasets}). In other words, the relevant activity type does not even occur. For such severe violations, the case duration (from the start to the end of the case) is used as the magnitude of violation, introducing a substantial number of outliers and resulting in a skewed data distribution. The hybrid approach, which focuses on regression tasks, is particularly sensitive to such imbalanced data distributions compared to the baseline (a classification approach) and the MTL approach.

Regarding compliance prediction performance, the MTL approach achieves results comparable to the baseline. However, the hybrid approach shows a significant decline in compliance prediction performance for the sepsis\_2 dataset, where the activity violation ratio is extremely high at 0.94 (cf. Tab.~\ref{tab:datasets}). By contrast, for the o2c\_1 dataset, the hybrid approach demonstrates a notable improvement, with an AUC value of 0.70. This improvement can be attributed to the synthetic dataset containing only basic attributes (case ID, activity, and timestamp) and temporal features extracted from the timestamps, which enhance the predictive performance for a regression task.

\begin{table}[htb!]
\centering
\caption{Comparison Performance for XGBoost}\label{tab:results_xgboost}
\begin{tabular}{c c c c c}
\hline
\multirow{2}{*}{Dataset} & \multirow{2}{*}{Approach} & Compliance & & Magnitude \\
\cline{3-3} \cline{5-5}
& & AUC & & MAE \\
\hline
\multirow{3}{*}{o2c\_1} & baseline & 0.50 & & 0.31 \\
                        & hybrid & \textbf{0.70} & & \textbf{0.21} \\
                        & multi-task & 0.50  & & 0.22 \\
\hline
\multirow{3}{*}{sepsis\_1}  & baseline & \textbf{0.57} & & 6.06 \\
                            & hybrid & 0.51 & & 8.41 \\
                            & multi-task & \textbf{0.57} & & \textbf{4.59} \\
\hline
\multirow{3}{*}{sepsis\_2}  & baseline & \textbf{0.87} & & 11.05 \\
                            & hybrid & 0.58 & & 13.11 \\
                            & multi-task & \textbf{0.87} & & \textbf{9.83} \\
\hline
\multirow{3}{*}{bpic2012w\_1} & baseline    & \textbf{0.56} & & 1.42 \\
                              & hybrid      & 0.51  & & 1.28 \\
                              & multi-task  & 0.55  & & \textbf{0.91} \\
\hline
\multirow{3}{*}{bpic2012w\_2} & baseline    & 0.53  & & 0.04 \\
                              & hybrid      & \textbf{0.55} & & \textbf{0.02} \\
                              & multi-task  & 0.52 & & \textbf{0.02} \\
\hline
\end{tabular}
\end{table}

\noindent\textbf{Att-Bi-LSTM.}
Table \ref{tab:results_lstm} summarizes the experimental results obtained with Att-Bi-LSTM. Similar to the magnitude prediction performance observed for XGBoost (cf. Tab.~\ref{tab:results_xgboost}), both hybrid and MTL approaches outperform the baseline. Notably, when Att-Bi-LSTM is used as the predictive model, our proposed approaches---particularly the MTL approach---achieve lower MAE values compared to XGBoost. This improvement stems from model adaptation in Att-Bi-LSTM (cf. Fig.~\ref{fig:att-bi-lstm_model_architecture}), where the magnitude prediction task leverages patterns captured by the compliance prediction task as additional input. 

The MTL approach slightly outperforms the baseline in compliance prediction. For the sepsis\_2 dataset, the hybrid approach exhibits a similar decline in compliance prediction performance as observed with XGBoost (cf. Tab.\ref{tab:results_xgboost}), due to the high activity violation ratio. Meanwhile, for the bpic2012w dataset, the hybrid approach underperforms other methods. This is likely due to the extremely low positive case ratios of 0.20 and 0.15 (cf. Tab.\ref{tab:datasets}), which result in imbalanced data distributions that negatively impact regression tasks.

\begin{table} 
\centering
\caption{Comparison Performance for Att-Bi-LSTM}\label{tab:results_lstm}
\begin{tabular}{c c c c c}
\hline
\multirow{2}{*}{Dataset} & \multirow{2}{*}{Approach} & Compliance & & Magnitude \\
\cline{3-3} \cline{5-5}
& & AUC & & MAE \\
\hline
\multirow{3}{*}{o2c\_1} & baseline      & \textbf{0.50}  & & 0.31 \\
                        & hybrid        & \textbf{0.50} & & \textbf{0.23}  \\
                        & multi-task    & \textbf{0.50}  & & \textbf{0.23}  \\
\hline
\multirow{3}{*}{sepsis\_1}  & baseline      & \textbf{0.60} & & 6.06 \\
                            & hybrid        & 0.58 & & 8.17 \\
                            & multi-task    & 0.56 & & \textbf{3.64}  \\
\hline
\multirow{3}{*}{sepsis\_2}  & baseline      & 0.86 & & 11.05 \\
                            & hybrid        & 0.55  & & 9.21 \\
                            & multi-task    & \textbf{0.87} & & \textbf{8.67} \\
\hline
\multirow{3}{*}{bpic2012w\_1}   & baseline      & 0.57  & & 1.42 \\
                                & hybrid        & 0.50 & & 1.34 \\
                                & multi-task    & \textbf{0.58} & & \textbf{0.85} \\
\hline
\multirow{3}{*}{bpic2012w\_2}   & baseline      & 0.55  & & 0.04 \\
                                & hybrid        & 0.50 & & 0.03 \\
                                & multi-task    & \textbf{0.56}  & & \textbf{0.02} \\
\hline
\end{tabular}
\end{table}

\noindent\textbf{Findings.}
In general, our proposed approaches outperform existing predicate prediction methods in quantifying the magnitude of violations, while maintaining comparable performance in compliance prediction. Specifically, when a deep learning architecture is used as the predictive model, the multi-task learning approach achieves superior results in both tasks. However, the performance of the hybrid approach is highly sensitive to data distribution. For instance, worse performance is observed on the sepsis datasets due to their skewed data distribution introduced by the extremely significant magnitudes of violations for deviant cases, which violate the temporal constraint on the control-flow perspective.

\section{Related Work}
\label{sec:rel_work} 

In this section, we analyze related work on predicting compliance states of ongoing process instances and quantifying the degree of compliance w.r.t. a given set of compliance constraints.

\noindent\textbf{Compliance Prediction.} \cite{DBLP:journals/tsc/Marquez-Chamorro18} classify predictive process monitoring (PPM) approaches into process-aware and non-process-aware approaches, both employing regression and classification methods to predict attributes in the future, such as next activity, next time, or the outcome of the case. As part of PPM, outcome-oriented PPM \cite{DBLP:journals/tkdd/TeinemaaDRM19,DBLP:conf/bpm/TeinemaaDMF16} and predicate prediction \cite{DBLP:conf/caise/MaggiFDG14,DBLP:journals/tsc/Francescomarino19} focus on compliance predictions of ongoing instances regarding a given set of constraints; the binary yes or no answer w.r.t. compliance is not sufficient for compliance management and requires the ability to quantify the degree of compliance.

\noindent\textbf{Compliance Degree.} \cite{DBLP:journals/ism/LuSG08} calculate a compliance distance to indicate the degree of match between the process model and the set of control rules at design time. They define control rules into four distinct classes according to the ideal semantics: ideal, sub-ideal, non-compliant and irrelevant states.
\cite{DBLP:conf/aicol/LamHK20} classify cases into full-, partial- and non-compliance and formulate a framework to detect and evaluate the degree of violations after process executions. However, the user-defined piece-wise mapping function cannot distinguish the extent of violations in a fine-grained manner. In \cite{DBLP:conf/edoc/MorrisonGK09}, a compliance scale model is created to measure the degree of compliance of completed process instances. Still, the mechanisms of c-semirings to classify process instances into $<Good, Bad>$ or $<0, .5, 1>$ are not accurate enough. \cite{Shamsaei2011IndicatorbasedPC} define a set of key performance indicators (KPIs) for each compliance rule and map the evaluation values of them to a compliance level from -100 to 100 taking into account values of target, threshold and worst. However, existing approaches provide various metrics for compliance degrees in different phases of process life cycle, but none of them target compliance prediction.

\section{Conclusion}\label{sec:conclusion}
This work proposes two predictive compliance monitoring approaches that extend state-of-the-art binary predicate prediction methods by enabling the quantification of violation magnitudes for ongoing process executions. The hybrid approach reformulates the binary classification problem into a regression task, focusing on quantifying violations in deviant cases, which implicitly aligns with the binary classification objective. By contrast, the multi-task learning (MTL) approach explicitly incorporates both compliance prediction and magnitude quantification tasks. This design is reasonable because these two tasks are closely related, and learning them together allows for feature sharing between tasks.

The evaluation demonstrates that our approaches---particularly the multi-task learning method---outperform existing predicate prediction methods in magnitude quantification while maintaining comparable performance in compliance prediction. The performance of both tasks improves when MTL leverages deep learning models, owing to the flexibility in adapting model architectures to better capture task-specific features. The ability to quantify the magnitude of violations is crucial in compliance monitoring, as it provides organizations with deeper insights into their operational performance and enables informed decision-makings to reduce or mitigate the risks of non-compliance.

However, our approaches have limitations. The performance of the hybrid approach is highly sensitive to data distribution, i.e., performing poorly on skewed datasets. This limitation is primarily due to the case labeling function applied to deviant cases, where the relevant event may not even occur. In future work, we aim to investigate methods for assigning the magnitudes of violation to deviant cases that breach temporal constraints in the control-flow dimension. Additionally, we plan to explore techniques to adapt the multi-task model architecture further, such as dynamically weighting each task's loss \cite{DBLP:conf/cvpr/KendallGC18}, to improve overall performance.

\begin{credits}
\subsubsection{\ackname} This work was funded by the Deutsche Forschungsgemeinschaft (DFG, German Research Foundation) -- project number 277991500.
\end{credits}
\bibliographystyle{splncs04}

\end{document}